# Multimodal Detection of COVID-19 Symptoms using Deep Learning & Probability-based Weighting of Modes

Meysam Effati[1]
meysam.effati@utoronto.ca

Yu-Chen Sun[1,2]
yuchen.sun@mail.utoronto.ca

Hani E. Naguib[2,3]
naguib@mie.utoronto.ca

Goldie Nejat[1,3], *Member, IEEE*
nejat@mie.utoronto.ca

[1]Autonomous Systems and Biomechatronics Laboratory (ASBLab)
[2]Toronto Smart Materials and Structures (TSMART)
Department of Mechanical and Industrial Engineering, University of Toronto
[3]Toronto Rehabilitation Institute

*Abstract*—The COVID-19 pandemic is one of the most challenging healthcare crises during the 21st century. As the virus continues to spread on a global scale, the majority of efforts have been on the development of vaccines and the mass immunization of the public. While the daily case numbers were following a decreasing trend, the emergent of new virus mutations and variants still pose a significant threat. As economies start recovering and societies start opening up with people going back into office buildings, schools, and malls, we still need to have the ability to detect and minimize the spread of COVID-19. Individuals with COVID-19 may show multiple symptoms such as cough, fever, and shortness of breath. Many of the existing detection techniques focus on symptoms having the same equal importance. However, it has been shown that some symptoms are more prevalent than others. In this paper, we present a multimodal method to predict COVID-19 by incorporating existing deep learning classifiers using convolutional neural networks and our novel probability-based weighting function that considers the prevalence of each symptom. The experiments were performed on an existing dataset with respect to the three considered modes of coughs, fever, and shortness of breath. The results show considerable improvements in detection of COVID-19 using our weighting function when compared to an equal weighting function.

## I. Introduction

The COVID-19 pandemic caused by the severe acute respiratory syndrome coronavirus 2 (SARS-CoV-2) can be considered one of the greatest medical challenges of our time. As of July 2021, there are over 192 million confirmed cases worldwide with more than 4.1 million confirmed deaths [1]. At the end of 2020 and early 2021, countries around the globe began mass vaccination and immunization programs to lower infection rates. Nevertheless, new virus mutation and variants with increased transmissibility have emerged, further extending the pandemic [2].

The symptoms for COVID-19 can range from asymptomatic to severe illness [3]. The virus is commonly transmitted by aerosol or droplets of contaminated fluid from an infected individual and enters a new host via the respiratory system [4]. As the virus starts to migrate down the respiratory pathway, the patient can start showing common symptoms such as fever, fatigue, and dry cough [3]. As these symptoms are similar to the common cold, the ability to accurately distinguish between the COVID-19 cases and other viruses is crucial for diagnosis and slowing down its spread. Currently, the real-time reverse transcription polymerase chain reaction (RT-PCR) test is one of the most accurate detection methods and is considered the gold standard for COVID-19 diagnosis in laboratory settings [5]. It consists of obtaining samples collected from the nose or throat to identify special markers [6]. Despite its reliability, RT-PCR has limitations when applied to mass populations. For example, specialized chemical reagents and facilities are required for the collected samples to be analyzed [6], [7]. Processing alone can take 4-6 hours, then the data needs to be translated by experts to the general public. The turnaround time from initial specimen collection to the release of the results may take up to 2-4 days depending on the available human resources and the testing capabilities [7], [8]. Therefore, a rapid screening process is needed to be deployed in the field, in order to provide immediate results and minimize the spread of the virus in public or crowded spaces.

It has been identified that early mild COVID-19 symptoms such as fever, coughing and/or shortness of breath can be used as features for preliminary diagnosis of COVID-19 [9]. In particular, automated detection of these symptoms using datasets have shown machine and deep learning algorithms have high classification rates for COVID versus non-COVID cases [10]–[13]. These methods can be categorized based on the detection techniques used: 1) imaging processing of X-ray or CT scans of the patients' chests [14], [15], 2) acoustic analysis of coughing/breathing sounds [16], [17], and 3) body temperature measurements using infrared thermometers [18].

The detection systems that consider more than one mode, consider them to be of equal importance. However, research has shown that certain symptoms are more prevalent in the detection of COVID-19 [19]. Both studies presented in [19], [20] found that fever, coughs, and shortness of breath are the most prevalent symptoms. As a result, the difference in the prevalence should be considered within multimodal classification, as the probability of each mode can influence the results of COVID screening.

In this paper we present a unique probability-based weighting function to incorporate the influence of multiple symptoms in determining COVID positivity. The weighting function can be integrated with classical or deep learning classifiers to determine the influence of each symptom mode in early detection of COVID-19 infections.

## II. RELATED WORK

COVID-19 detection methods can be categorized into two groups: 1) unimodal detection methods, 2) multi-modal methods. We discuss the symptoms and techniques used within these categories below.

### A. Unimodal Automated Methods for COVID-19 Symptom Detection

At the early stages of the pandemic, many studies focused on unimodal automated methods for COVID-19 detection, in particular on X-ray [14], CT [15], or ultrasound [21] images of patients' chests to identify anomalies [25].

Despite the high accuracy of AI enhanced imaging identification, these diagnosis methods are not portable and must be performed in hospitals or medical centers where the equipment is available. Furthermore, specially trained technicians are needed in order to operate these systems, which further restricts the testing capability for identifying COVID-19 in the field. Therefore, researchers have been investigating the use of more widely available signals for development of early detection of COVID symptoms [26]. These include, for example, vocal signals such as coughing [22], [27]–[29] or breathing [23], [30], [31].

Coughing is a very common and natural physiological behavior which is associated with several viral respiratory infections including COVID-19 [32], [33]. Depending on the infected and the irritant locations within the respiratory system, the coughing sound produced by different respiratory infections have distinct features [34]–[39]. Recent studies have found that COVID-19 infects the respiratory system in a distinct way [27], [40]–[42]. By comparing the CT analysis of COVID-19 infected patients with non-COVID pneumonia, the COVID-19 patients are more likely to develop a peripheral distribution, ground-glass opacity, vascular thickening, and reverse halo sign [16], [40].

Based on the difference of changes in the respiratory system, a COVID-19 patient would likely produce a distinct coughing sound that can be identified by learning algorithms [16]. For example, in [28], Convolutional Neural Networks (CNN) were used to identify people with COVID through forced coughing as input signals. When processing the raw coughing signal, four distinct biomarkers including muscular degradation, changes in vocal cords, changes in sentiment/mood, and changes in the lungs/respiratory tract were utilized as part of the detection criteria.

Another common symptom of COVID-19 infection is changes in the breathing rhythm or shortness of breath. In [23], frequency-modulated continuous wave signals were classified using a XGBoost classifier and a Mel-frequency cepstral coefficient (MFCC) feature extractor. Five different breathing patterns were analyzed: normal breathing, deep/quick breathing, deep breathing, quick breathing, and holding the breath.

Despite the fact that fever is the most common symptoms of COVID-19 [20], a unimodal temperature measurement is not sufficient to determine whether a person is infected by the virus, as fever alone is a very common symptom of many illnesses [43]. Furthermore, fever can be less common in younger age groups that have been diagnosed with mild or moderate COVID [44]. As a result, temperature can only be utilized in combination with other modes (symptoms) in multimodal analysis.

### B. Multi-modal Automated Methods for COVID-19 Detection

Since COVID-19 patients are usually presented with multiple symptoms [24], multi-modal deep learning classification can be used to improve the accuracy of COVID-19 diagnoses [17]. For example, in [30], a COVID-19 Identification ResNet (CIdeR) classifier used cough and breathing together to determine the probability of COVID-positivity. In [17], more modes were utilized. Namely, Logistic Regression (LR) and Support Vector Machines (SVM) were used to classify breathing, coughing, and speech signals separately and a Decision Tree classifier was used to classify a group of other symptoms such as fatigue, muscle pain, and loss of smell. The outputs of each classifier were prediction scores and were all given equal weights to determine the overall probability of COVID.

However, when investigating clinically obtained statistical datasets for COVID-19, it has been found that varying symptoms have different prevalence, as seen in the MIT dataset [22]. In this paper we introduce a unique weighting system that can utilize multiple modes (non-fixed parameters) to determine the influence of each individual symptom for COVID positivity.

## III. MULTI-MODAL DETECTION USING PROBABILITY-BASED WEIGHTING OF MODES

Our proposed methodology consists of using multiple symptoms as input to our multi-modal COVID detection architecture to determine the probability of COVID positivity based on the prevalence of $n$ number of symptoms, $s_i (i = 1, \ldots, n)$. The overall methodology is shown in Fig. 1. Different mode inputs including cough, breathing, speech, and images can be classified with corresponding machine learning (ML) and deep learning (DL) classifiers. In addition, other symptoms as fatigue, loss of smell, and fever can be utilized as inputs in the binary classifiers, such as decision trees. The output of these classifiers are label indicators $(I_{S_i} \in \{0, 1\})$, and given as inputs to the weighted function, $f_W$:

$$f_w = \sum_{i=1}^{n} I_{S_i} w_{s_i}. \qquad (1)$$

where $I_{S_i}$ ($i \in \{1, \ldots, n\}$) which is the output of each classifier is defined as follows:

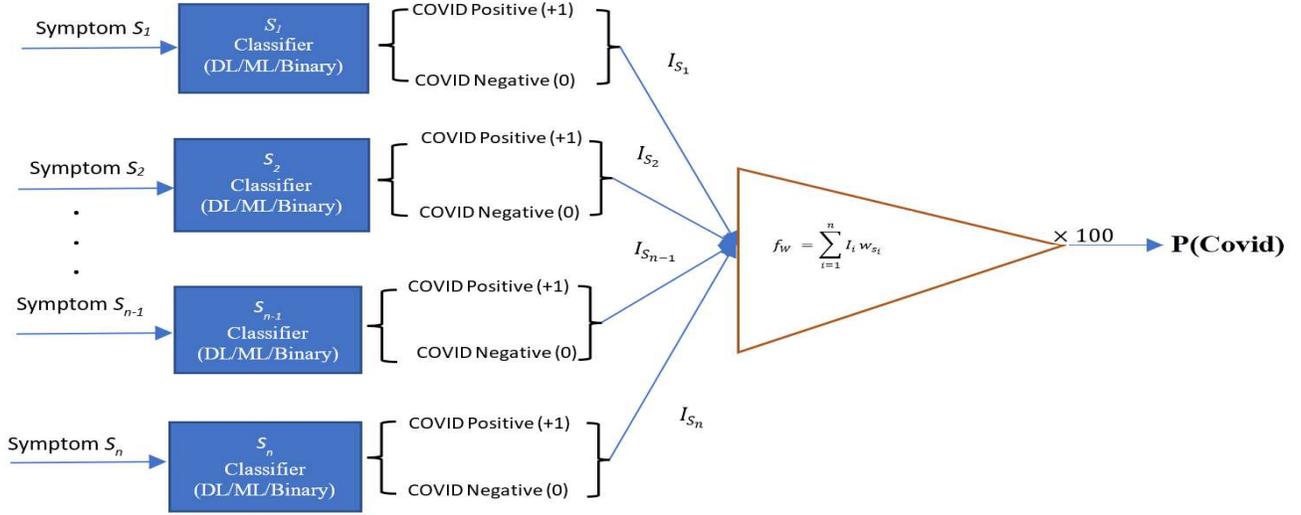

Fig. 1. Symptom classification and probability-based weighted function architecture for Covid-19 detection of *n* modes

$$I_{S_i} = \begin{cases} 1, & \text{for the classified COVID} - \text{positive symptom} \\ 0, & \text{for the classified COVID} - \text{negative symptom} \end{cases}$$

and the weights are obtained as:

$$w_{S_i} = \frac{p_i}{\sum_{j=1}^{n} p_j}, \quad (2)$$

where $j \in \{1, \ldots, n\}$. Also, $p_i$ and $p_j$ are the prevalence of each symptom. These values ($p_i$ and $p_j$) represent the probability that a person who tested positive for COVID-19 has the considered symptom. This weighting function incorporates the influence of each individual symptom. The output of $f_W$ (x100), provides the probability of COVID-positivity.

IV. TRAINING

We used coughs, breathing and fever as COVID symptoms to train our architecture. The CIdeR CNN network presented in [30] was used to train two separate networks using the Cambridge dataset [45]. The Cambridge dataset includes 459 crowdsourced labeled cough and breathing audio recordings in the webm format [45]. All the information is self-declared. We use a 70%-30% split for training and testing. We obtained similar AUCs as reported in [30] for our CNN networks. A decision tree classifier is used to classify fever.

In order to obtain $p_i$ and $p_j$ for each of the three symptoms for our weighting function, the MIT dataset [22] is utilized. Namely, the dataset includes COVID-19 data from over 160 clinical studies of different countries, including China, USA, Japan, Singapore, and Italy. This statistical dataset contains the *study population* and *symptoms prevalence* (in percentage) for the related study population. The symptoms include cough, shortness of breath, fever, fatigue, etc. Herein, we use the average of prevalence for each symptom in the dataset. Hence, $p_C, p_B,$ and $p_F$, which are the prevalence for cough, breathing, and fever are obtained: $(p_C, p_B, p_F) = (99.19, 59.35, 29.09)$, where all the values are in percentages. Eq. (2) is utilized to obtain the weights for each of the symptoms $(w_C, w_B, w_F) = (0.32, 0.15, 0.53)$. Hence, the following relation for the weighting function is obtained:

$$f_W = 0.32 I_C + 0.15 I_B + 0.53 I_F. \quad (3)$$

V. EXPERIMENTS

The objective of the experiments is to investigate the performance of our weighting function ($f_W$) with respect to the COVID-positivity rate of the test data. In particular, we compared our probability-based weighting function ($f_W$) with that of an equal weighting function [17]:

$$f_r = (1/3) I_C + (1/3) I_B + (1/3) I_F \quad (4)$$

We investigated four different cases:

**1) Case 1: False Negative for Breathing-** In this case, cough recordings and fever measurements are correctly classified as COVID-positive ($I_C = 1$ and $I_F = 1$) by the CIdeR and decision tree classifiers. However, breathing is incorrectly classified as COVID-negative by the CIdeR classifier ($I_B = 0$). This case happens for 7 inputs.

**2) Case 2: False Positive for Breathing-** Cough recordings are correctly classified as COVID-negative ($I_C = 0$) and breathing is incorrectly classified as COVID-positive ($I_B = 1$). Users did not have a fever, i.e., ($I_F$) is considered 0. This case occurs for 198 inputs, which is a 63% of all inputs.

**3) Case 3: False Negative for both Cough and Breathing-** Both cough and breathing recordings are incorrectly classified as COVID-negative ($I_B = 0$ and $I_C = 0$) by the CIderR classifiers. $I_F$ is 1 as the users have a fever. This case occurs for 105 inputs.

TABLE I. Comparison results for $f_W$ and $f_r$ weighting functions for the different cases.

| Case# | Case Type | Number of Occurrences for the Case | Indicator Labels (output of Classifiers) | $f_W$ | $f_r$ |
|---|---|---|---|---|---|
| 1 | False Negative for Breathing | 7 | $I_C = 1$<br>$I_B = 0$<br>$I_F = 1$ | 0.85 | 0.67 |
| 2 | False Positive for Breathing | 198 | $I_C = 0$<br>$I_B = 1$<br>$I_F = 0$ | 0.15 | 0.33 |
| 3 | False Negative for Cough and Breathing | 105 | $I_C = 0$<br>$I_B = 0$<br>$I_F = 1$ | 0.53 | 0.33 |
| 4 | False Positive for Cough and Breathing | 6 | $I_C = 1$<br>$I_B = 1$<br>$I_F = 0$ | 0.47 | 0.67 |

**4) Case 4: False Positive for both Cough and Breathing-** Both cough and breathing recordings are incorrectly classified as COVID-positive ($I_B = 1$ and $I_C = 1$) by the CIderR classifiers. $I_F$ is considered 0 by the decision tree classifier as the users do not have a fever.

*A. Results*

The results are presented in Table I. The relative differences between $f_W$ and $f_r$ are determined to be: 26.9% for **Case 1**; -54.5% for **Case 2**; 60.6% for **Case 3**; and -29.9% for **Case 4**, respectively. For **Cases 1**, and **3**, our probability-based weighting function increases the detected COVID probability rate even when the classifiers incorrectly classify the input symptoms as negative- *therefore, it can reduce the impact of false negative results*. Furthermore, for **Cases 2** and **4**, our weighting function helps improve the detection of the COVID-negativity rate when input symptoms are incorrectly classified as COVID positive- *namely it can reduce the false positive rate for COVID detection*. It is worth noting that for **Case 2**, this occurred for 198 input sets which is a large portion (63%) of the input symptom sets in the testing dataset.

Overall, the percent improvement of using $f_W$ in comparison with $f_r$, normalized across the four cases considered herein is 55.4%. Namely, the results show that when the multiple symptoms of cough, breathing and fever are considered, a probability-based weighting function can help improve the early detection of COVID.

CONCLUSIONS

A novel clinical weighting system is proposed to detect COVID-19 using multi-modal symptoms. The multimodal system is used to predict COVID-19 positivity by incorporating learning/binary classifiers and our unique probability-based weighting function that considers the prevalence of each symptom. A comparison study of our proposed function with an equal weighting function showed that we are able to increase the probability of early detection of COVID-19 by an average of 55.4%. Future work will include the inclusion of additional modes for early COVID-19 detection and the testing of our architecture in the field.

ACKNOWLEDGMENT

The authors acknowledge funding from AGE-WELL Inc. and the Canada Research Chairs (CRC) Program. We would also like to thank the University of Cambridge for sharing the COVID-19 dataset.